\documentclass[12pt]{colt2016}

\usepackage{times}
\usepackage{dsfont}
\usepackage{soul,color}
\usepackage{graphicx,float,wrapfig}
\usepackage[usenames,dvipsnames]{pstricks}
\usepackage{pst-grad} 
\usepackage{pst-plot} 

\newtheorem{theo}{Theorem}[section]

\newtheorem{conj}[theo]{Conjecture}
\newtheorem{defi}[theo]{Definition}
\newtheorem{rem}[theo]{Remark}

\newcommand{\topic}[1]{\vspace{0.2cm}\noindent{\bf {#1}:}}

\newcommand{\N}{\mathbb{N}}

\newcommand{\bestarm}{\textsc{Best-$1$-Arm}}

\newcommand{\CORRECT}{$\delta$-correct}

\newcommand{\eat}[1]{}

\newcommand{\arm}{A}
\newcommand{\distr}{\mathcal{D}}
\newcommand{\alg}{\mathbb{A}}

\newcommand{\Gap}[1]{\Delta_{[#1]}}

\title{Open Problem: Best Arm Identification:  Almost Instance-Wise Optimality and 
	the Gap Entropy Conjecture}
	
\coltauthor{\Name{Lijie Chen}\thanks{ Institute for Interdisciplinary Information Sciences (IIIS), Tsinghua University, Beijing, China.
		Research supported in part by the National Basic
		Research Program of China grants 2015CB358700,
		2011CBA00300, 2011CBA00301, and the National NSFC
		grants 61033001, 61361136003.
	} \Email{chenlj13@mails.tsinghua.edu.cn}\\
 \addr Institute for Interdisciplinary Information Sciences (IIIS),
Tsinghua University, China.
 \AND
 \Name{Jian Li}\footnotemark[1] \Email{lijian83@mail.tsinghua.edu.cn}\\
 \addr Institute for Interdisciplinary Information Sciences (IIIS),
Tsinghua University, China.
 }

\begin{document}
	\maketitle 	
\vspace{-1cm} 
\begin{abstract}
	The best arm identification problem (\bestarm) is the most basic pure exploration 
	problem in stochastic multi-armed bandits. 
	The problem has a long history and attracted significant attention 
	for the last decade. 
	However, we do not yet have a complete understanding of the
	optimal sample complexity of the problem:
	The state-of-the-art algorithms achieve a sample complexity 
	of $O(\sum_{i=2}^{n} \Delta_{i}^{-2}(\ln\delta^{-1} + \ln\ln\Delta_i^{-1}))$
	($\Delta_{i}$ is the difference between the largest mean and the $i^{th}$ mean), while the best known lower bound is $\Omega(\sum_{i=2}^{n} \Delta_{i}^{-2}\ln\delta^{-1})$ for general instances and $\Omega(\Delta^{-2} \ln\ln \Delta^{-1})$ for the two-arm instances. 
	We propose to study the instance-wise optimality for the  \bestarm\ problem.
	Previous work has proved that it is impossible to have an instance optimal
	algorithm for the 2-arm problem. 
	However, we conjecture that modulo the additive term
	$\Omega(\Delta_2^{-2} \ln\ln \Delta_2^{-1})$ (which is an upper bound and worst case lower bound for the 2-arm problem), 
	there is an instance optimal algorithm for \bestarm.  
	Moreover, we introduce a new quantity, called the gap entropy for a best-arm problem instance, and conjecture that it is the instance-wise lower bound. 
	Hence, resolving this conjecture would provide a final answer 
	to the old and basic problem.
\end{abstract}

\vspace{-0.4cm} 
\section{Introduction}

In the \bestarm\ problem, we are given $n$ stochastic arms $\arm_1,\ldots, \arm_n$.
	 The $i^{th}$ arm $\arm_i$ has a reward distribution $\distr_i$ with an unknown mean $\mu_{i}\in [0,1]$. 
	 We assume that all reward distributions are Gaussian distributions with variance 1.   
	 Upon each play of $\arm_i$,
	 we can get a reward value sampled i.i.d. from $\distr_i$. Our goal is to identify the arm with largest mean using as few samples as possible.
	 We assume here that the largest mean is strictly larger than the second largest (i.e., $\mu_{[1]}>\mu_{[2]}$)
	 to ensure the uniqueness of the solution,
	 where $\mu_{[i]}$ denotes
	 the $i^{th}$ largest mean. The problem is also called the {\em pure exploration problem} in the stochastic multi-armed bandit literature.
	 
We say an algorithm $\alg$ is $\delta$-correct for \bestarm, if it outputs the correct answer on {\em any} instance with probability at  $1-\delta$, and we use $T_{\alg}(I)$ to denote the expected number of total samples taken by algorithm $\alg$ on instance $I$. We also define the gap of $i^{th}$ arm, $\Gap{i} := \mu_{[1]} - \mu_{[i]}$.

\vspace{-0.4cm} 
\section{Background}

	During the last decade,  
	the \bestarm\ problem and its optimal sample complexity
	have attracted significant attention.
	We only mention a small subset that are most relevant to us. 
	The current best lower bound is due to \cite{mannor2004sample},
	who showed that for any \CORRECT\ algorithm for \bestarm, 
	it requires $\Omega\left(\sum\nolimits_{i=2}^{n} \Gap{i}^{-2} \ln\delta^{-1}\right)$ (referred to as the MT lower bound from now on) 
	samples in expectation for any instance.
	We note that the MT lower bound is an {\em instance-wise lower bound}, 
	i.e., any \bestarm\ instance requires the stated number of samples.
	On the other hand, the current published best known upper bound is
	$O\left(\sum\nolimits_{i=2}^{n} \Gap{i}^{-2} \left(\ln\ln\Gap{i}^{-1}+\ln\delta^{-1}\right)\right)$,
	due to \cite{karnin2013almost}.
	\cite{jamieson2014lil} obtained a UCB-type algorithm (called lil'UCB), which achieves
	the same sample complexity. 
	We refer the above bound as the KKS-JMNS bound.
	Back in 1964, \cite{farrell1964asymptotic} provided an $\Omega(\Delta^{-2}\ln\ln\Delta_2^{-1})$ lower bound for the two-arm cases
	(which matches the KKS-JMNS bound for two arms).

	\eat{	
	In fact, some researchers have believed that the KKS upper bound is already optimal,
	since it matches \cite{farrell1964asymptotic}'s $\Omega(\Delta^{-2}\ln\ln\Delta^{-1})$ lower bound for two arms. 
		Both \cite{jamieson2014lil} and \cite{jamieson2014best} explicitly referred the upper bound as ``optimal''.
		In \cite{jamieson2014lil}, it (erroneously) states that ``{\em The procedure cannot be improved in the sense that the number
		of samples required to identify the best arm is within a constant factor of a lower bound based on the law of the iterated logarithm (LIL)}''.
	}
	
	Very recently, in an unpublished manuscript (\cite{chen2015optimal}), the authors
	obtained improved lower and upper bounds for \bestarm.
	The work lead the authors to make an intriguing conjecture which we detail in
	the next section.
	We will also state the improved bounds and their connection to the conjecture in
	more details. 

\vspace{-0.4cm} 	
\section{Open Problem: Almost Instance Optimality and the Gap Entropy Conjecture}

\newcommand{\ordlow} {\mathcal{L}}
\newcommand{\arment} {\mathsf{Ent}}

We propose to study \bestarm\ from the perspective of 
instance optimality, the ultimate notion of optimality  
(see e.g., \cite{fagin2003optimal,afshani2009instance}).

For the 2-arm cases, the KKS-JMNS bound
$O(\Delta^{-2}\ln\ln\Delta_2^{-1})$ is an upper bound for every instance,
and the Farrell lower bound
$\Omega(\Delta^{-2}\ln\ln\Delta_2^{-1})$ is a lower bound for the worst case instances.
As we observed in (\cite{chen2015optimal}), it is impossible to obtain an instance optimal algorithm even for the 2-arm cases.
\eat{
However, it is not hard to see that $\ordlow(I_{\Delta},\delta) \le O(\Delta^{-2} \ln \delta^{-1})$, since we do have a \CORRECT\ algorithm for \bestarm\ with complexity $O(\Delta^{-2}\ln\delta^{-1})$ on the instance $I_{\Delta}$'s. \footnote{Which can be constructed by using Corollary 3.3 in \cite{chen2015optimal}.} So even for two-arm case, we do not have algorithm with the instance-wise optimal complexity $O(\ordlow(I,\delta))$.
}
While the observation has ruled out any hope of an instance optimal algorithm for \bestarm, however, as we will see, it is still possible 
to obtain very satisfiable answer in terms of instance optimality.

Now, we formally define what is an instance-wise lower bound. Clearly, two arm instances differ only by a permutation of arms should be considered as the same instance. 
Inspired by \cite{afshani2009instance}, we give the following natural definition.

\begin{defi}(Order-Oblivious Instance-wise Lower Bound)
	
	Given a \bestarm\ instance $I$ and a confidence level $\delta$, we define
\vspace{-0.2cm} 
	$$
	\ordlow(I,\delta) := \inf_{\alg: \alg \text{ is } 
	\delta\text{-correct for \bestarm}} \,\,\frac{1}{n!} \cdot \sum\nolimits_{\pi \in \mathrm{Sym}(n)} T_{\alg}(\pi \circ I),
	$$\vspace{-0.2cm} 
	where the summation is over all $n!$ permutations of $\{1,\ldots, n\}$. 
\end{defi}
The MT lower bound immediately implies that $\ordlow(I,\delta) = \Omega(\sum\nolimits_{i=2}^{n} \Gap{i}^{-2}\ln\delta^{-1})$.

We conjecture that
the two-arm instance is the {\em only} obstruction toward an instance-wise optimal algorithm. More precisely, we have the following conjecture.
\begin{conj}
	\label{conj:optimal}
	There is an algorithm for \bestarm\ with sample complexity
\vspace{-0.2cm}
	$$
	O(\ordlow(I,\delta) + \Delta_2^{-2}\ln\ln\Delta_2^{-1}),
	$$ for any instance $I$ and $\delta < 0.1$. 
	And we say such an algorithm is almost instance-wise optimal for \bestarm.
\end{conj}
In the light of the discussion for the 2-arm cases, 
there must be a gap between the sample complexity of a \CORRECT\ algorithm and $\ordlow(I,\delta)$, and Conjecture~\ref{conj:optimal} states
that the gap can be as small as an {\em additive factor} $\Delta_2^{-2}\ln\ln\Delta_2^{-1}$, which is all we need to find out the best arm
from the top-2 arms, and is an inevitable gap even for the 2-arm instances. 

Moreover, we provide an explicit formula for $\ordlow(I,\delta)$.
Interestingly, the formula involves 
an entropy term (similar entropy terms also appear in \cite{afshani2009instance}
for completely different problems). 
We define the entropy term first.

\begin{defi}
Given a \bestarm\ instance $I$, 
let 
\vspace{-0.2cm}
$$
G_k = \{i \in [2,n] \mid  2^{-k} \le \Gap{u} < 2^{-k+1} \},\quad
H_k = \sum\nolimits_{i \in G_k} \Gap{i}^{-2},
\quad\text{ and }\quad
p_k = H_k/\sum\nolimits_j H_j.
$$
We can view $\{p_k\}$
as a discrete probability distribution.
We define the following quantity as the {\em gap entropy} for the instance $I$
$$
\arment(I) = \sum\nolimits_{G_k \ne \emptyset} p_k \log p_k^{-1}.\footnote{Note that it is exactly the Shannon 
entropy for the distribution defined by $\{p_k\}$.}
$$
\end{defi}
\vspace{-0.5cm}
\begin{rem}
	We choose to partition the arms based on the powers of $2$. 
	There is nothing special about 2 and replacing it
	by any other constant only changes $\arment(I)$
	by a constant factor.	
\end{rem}
\vspace{-0.2cm}
Then we formally state our conjecture.
\begin{conj}\label{conj:opt-ent}
	For any \bestarm\ instance $I$ and $\delta < 0.1$, we have
	\vspace{-0.2cm}
	$$
	\ordlow(I,\delta) = \Theta\left(\sum\nolimits_{i=2}^{n} \Gap{i}^{-2}\cdot\left(\ln\delta^{-1} + \arment(I)\right)\right).
	$$
\end{conj}
\vspace{-0.2cm}
In the next section, we will try to motivate the term $\arment(I)$
and explain the reasons that lead us to make the above conjecture.
\vspace{-0.4cm} 
\section{Motivation and Current Progress}
In our recent work (\cite{chen2015optimal}), 
we provide an algorithm with the following sample complexity:
\vspace{-0.2cm}
\begin{align}
	\label{eq:ub}
O\Big(
\Gap{2}^{-2}\ln\ln \Gap{2}^{-1}+
\sum\nolimits_{i=2}^{n} \Gap{i}^{-2} \ln\delta^{-1}+\sum\nolimits_{i=2}^{n} \Gap{i}^{-2}\ln\ln \min(n,\Gap{i}^{-1}) 
\Big).
\end{align}
Furthermore, the algorithm achieves a sample complexity of
\vspace{-0.2cm}
\begin{align}
	\label{eq:ub2}
O\left( \Gap{2}^{-2}\ln\ln \Gap{2}^{-1}+\sum\nolimits_{i=2}^{n} \Gap{i}^{-2} \ln\delta^{-1}\right),
\end{align}
for clustered instances
(We say an instance is clustered if 
	the number of nonempty $G_k$s is bounded by a constant).

Our new upper bounds \eqref{eq:ub} and \eqref{eq:ub2}
match our conjectured gap entropy lower bound in 
two extreme cases. 
On one extreme, the maximum value $\arment(I)$ can get is 
$O(\ln\ln n)$.
This can be achieved by instances in which there are $\log n$ nonempty groups $G_i$ and
they have almost the same weight $H_i$.
Hence, \eqref{eq:ub} is optimal for such instances. 
On the other extreme where
there is only a constant number of nonempty groups
(i.e., the instance is clustered),
$\arment(I)=O(1)$, and our algorithm can achieve almost instance optimality (without relying on the Conjecture~\ref{conj:opt-ent}, due to the MT lower bound) in this case.

Besides the fact that our algorithm can achieve
optimal results for both extreme cases,
we have more reasons to believe why 
$\arment(I)$ should enter the picture.
	
\topic{Upper Bounds}

First, we motivate the gap entropy $\arment$ 
from the algorithmic side.
Consider an elimination-based algorithm 
(such as \cite{karnin2013almost} or our algorithm).
We must ensure that the best arm is not eliminated 
in any round. 
Recall that in the $r^{th}$ round, we want to eliminate arms with gap $\Delta_r = \Theta(2^{-r})$, which is done by obtaining an approximation of the best arm, then take $O(\Delta_r^{-2} \ln \delta_r^{-1})$ samples from each arm and eliminate the arms with smaller empirical means. 
Roughly speaking, we need to assign the failure probability $\delta_r$ carefully to each round
(by union bound, we need $\sum_r\delta_r\leq \delta$).
The algorithm in \cite{karnin2013almost}
used $\delta_r = O(\delta \cdot r^{-2})$, and we used a better way to assign $\delta_r$. 
Indeed, if one can assign
$\delta_r$'s optimally (i.e., minimize 
$\sum_{r} H_r \ln \delta_r^{-1}$ subject to $\sum_{r} \delta_r \le \delta$),
one could achieve the entropy bound 
$\sum_r H_r \cdot (\ln\delta^{-1} + \arment(I))$
(by letting $\delta_r = \delta H_r/\sum_i H_i $).
Of course, this does not lead to
an algorithm directly,
as we do not know $H_i$s in advance. 

Using our techniques, 
we can estimate the values $H_r$'s when we enter the $r^{th}$ elimination stage. 
The only obstacle for implementing the above idea of assigning $\delta_r$'s optimally is that we do not know $\sum_{r} H_r$ initially. 
We believe the difficulty can be overcome by additional new algorithmic ideas. 

\topic{Lower Bounds}

In \cite{chen2015optimal}, we prove the following lower bound, improving the 
MT lower bound.
		\begin{theo}(Theorem 1.6 in \cite{chen2015optimal})\label{thm:hard-case-exists}
			There exist constants $c,c_1 > 0$ and $N \in \N$ such that, for any $\delta < 0.005$ and any \CORRECT\ algorithm $\alg$, 
			and any $n \ge N$, there exists an $n$ arms instance $I$ such that 
			$T_{\alg}[I] \ge c \cdot \sum_{i=2}^{n} \Gap{i}^{-1} \ln\ln n$. 
			Furthermore, $\Gap{2}^{-2} \ln\ln \Gap{2}^{-1} < \frac{c_1}{\ln n}\cdot \sum_{i=2}^{n} \Gap{i}^{-1} \ln\ln n$.
		\end{theo}

In fact, in the lower bound instances, there are $\log n$ nonempty groups $G_i$ and
they have almost the same weight $H_i$ (hence, $\arment(I) = \Theta(\ln\ln n)$).
Combining with the MT lower bound, we have covered the two extreme ends 
of Conjecture~\ref{conj:opt-ent}.

Moreover, it is possible to extend 
our current technique to construct many instances $I_S$ such that any algorithm 
$\alg$ requires at least $\Omega( H(I_S) \cdot \arment(I_S))$ samples. 
This strongly suggests $\Omega( H(I) \cdot \arment(I))$ is the right lower bound. 
However, a complete resolution of Conjecture~\ref{conj:opt-ent} seems to require
new techniques.
\vspace{-0.4cm} 
	\bibliographystyle{abbrv}
	\bibliography{team} 

\end{document}